\documentclass[11pt, a4paper, logo, copyright, nonumbering]{deepseek}
\usepackage[numbers, sort&compress]{natbib}
\usepackage{dblfloatfix}
\usepackage{ulem}
\usepackage{caption}
\usepackage{dramatist}
\usepackage{xspace}
\usepackage{pifont}
\usepackage{multirow}
\usepackage{tcolorbox}
\usepackage{xltabular}
\usepackage{longtable}
\usepackage{hyperref}
\usepackage{amsfonts}
\usepackage{amsmath}
\usepackage{amssymb}
\usepackage{lineno}
\usepackage{wrapfig}  
\usepackage{booktabs} 
\usepackage[bottom]{footmisc}

\usepackage{CJKutf8}
\usepackage{subfigure}
\usepackage{setspace}
\usepackage{xcolor}
\usepackage{arydshln}
\definecolor{linkcolor}{rgb}{0.956,0.298,0.235}
\definecolor{citecolor}{HTML}{1976D2}
\hypersetup{
    colorlinks=true,    
    linkcolor=citecolor,     
    filecolor=magenta,  
    urlcolor=cyan,      
    citecolor=citecolor,    
}

\newcommand{\ours}{JanusFlow}
\newcommand{\D}{\mathcal{D}}
\renewcommand{\d}{{\rm{d}}}
\newcommand{\p}{{\rm{P}}}
\newcommand{\E}{{\mathbb{E}}}

\makeatletter
\def\@BTrule[#1]{%
  \ifx\longtable\undefined
    \let\@BTswitch\@BTnormal
  \else\ifx\hline\LT@hline
    \nobreak
    \let\@BTswitch\@BLTrule
  \else
     \let\@BTswitch\@BTnormal
  \fi\fi
  \global\@thisrulewidth=#1\relax
  \ifnum\@thisruleclass=\tw@\vskip\@aboverulesep\else
  \ifnum\@lastruleclass=\z@\vskip\@aboverulesep\else
  \ifnum\@lastruleclass=\@ne\vskip\doublerulesep\fi\fi\fi
  \@BTswitch}
\makeatother

\addto\extrasenglish{
}

 {\begin{list}{}%
         {\setlength{\leftmargin}{#1}}%
         \item[]%
 }
 {\end{list}}

\reportnumber{001} 

\renewcommand{\today}{}

\title{\centering \ours: Harmonizing Autoregression and Rectified Flow for Unified Multimodal Understanding and Generation}

\author[*]{

\small
Yiyang Ma$^{1, 2}$ \quad Xingchao Liu$^{1, \dag}$ \quad Xiaokang Chen$^{1, \dag}$ \quad Wen Liu$^{1, \dag}$ \quad Chengyue Wu$^{1, 3}$ \quad Zhiyu Wu$^{1, 2}$ \quad
Zizheng Pan$^1$ \quad Zhenda Xie$^1$ \quad Haowei Zhang$^1$ \quad Xingkai Yu$^1$ \quad Liang Zhao$^1$ \quad Yisong Wang$^{1, 4}$ \quad 
Jiaying Liu$^2$ \quad Chong Ruan$^{1, \ddagger}$
\\
\vspace{4mm}

\small
$^1$DeepSeek-AI \quad $^2$Peking University \quad $^3$The University of Hong Kong \quad $^4$Tsinghua University \\

\small
$^\dag$Equal contribution, $^\ddagger$Corresponding author \\
\small
Project Page: \url{https://github.com/deepseek-ai/Janus}
}

\renewcommand{\phi}{\varphi}

\renewcommand{\geq}{\geqslant}

\renewcommand{\epsilon}{\varepsilon}
\renewcommand{\imath}{\mathrm{i}}

\newlength{\restsubwidth}
\newlength{\restsubheight}
\newlength{\restsubmoreheight}
\newcommand{\rest}[2]{%
        \settowidth{\restsubwidth}{\ensuremath{#2}}
        \settoheight{\restsubheight}{\ensuremath{{}_{#2}}}
        \ensuremath{{#1\hskip 0.5pt}_{\vrule\kern2pt\parbox[b][%
        4pt][b]{\the\restsubwidth}{%
                        \ensuremath{{}_{#2}}}}}
        }

\newcommand{\eg}{\textit{e}.\textit{g}., }

\begin{abstract}
We present \textbf{\ours}, a powerful framework that unifies image understanding and generation in a single model.
\ours~introduces a minimalist architecture that integrates autoregressive language models with rectified flow, a state-of-the-art method in generative modeling.
Our key finding demonstrates that rectified flow can be straightforwardly trained within the large language model framework, eliminating the need for complex architectural modifications.
To further improve the performance of our unified model, we adopt two key strategies: (i) decoupling the understanding and generation encoders, and (ii) aligning their representations during unified training.
Extensive experiments show that \ours~achieves comparable or superior performance to specialized models in their respective domains, while significantly outperforming existing unified approaches across standard benchmarks.
This work represents a step toward more efficient and versatile vision-language models.
\end{abstract}

\begin{document}
\begin{CJK*}{UTF8}{gbsn}

\maketitle

\section{Introduction}

\begin{figure}[t]
\begin{center}
    \subfigure[Benchmark Performances.]
    {
        \includegraphics[width=0.41\linewidth]{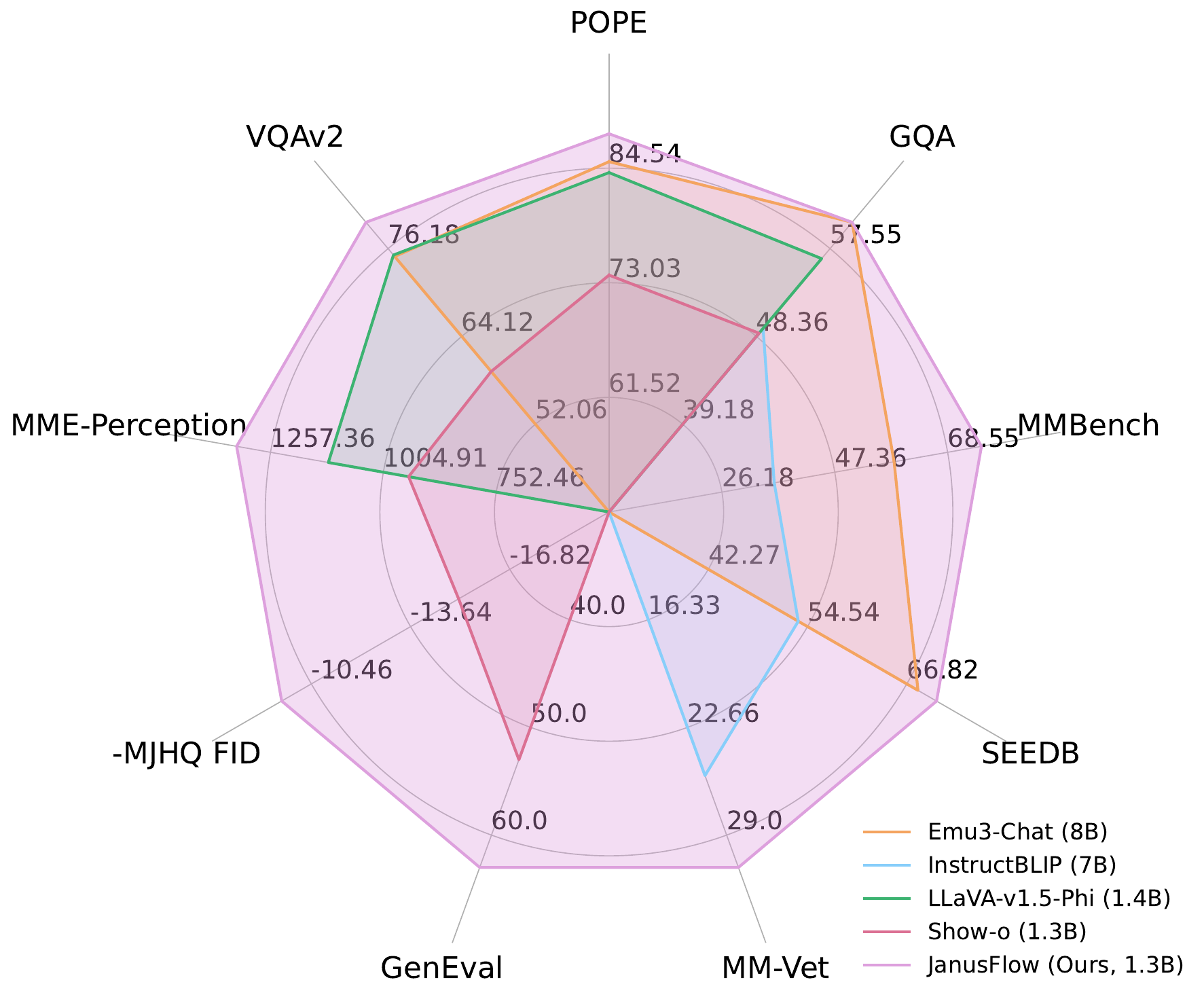}
        \label{fig:teaser_understanding}
    }
    \hfill
    \subfigure[Visual Generation Results.]
    {
        \includegraphics[width=0.55\linewidth]{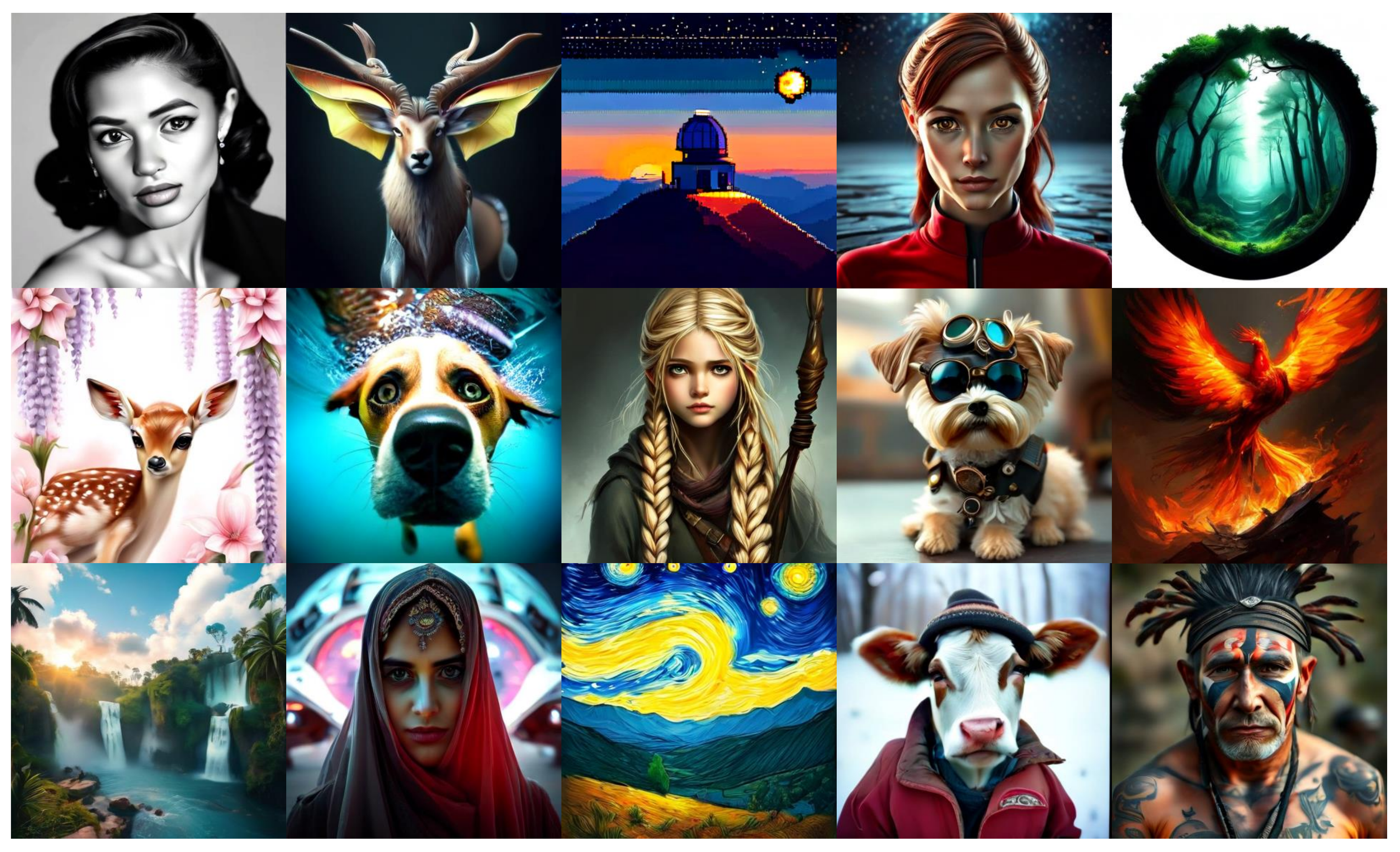}
        \label{fig:teaser_generation}
    }
\end{center}
\caption{\small
\textbf{
Multimodal understanding and image generation with \ours}. \ours~surpasses the state-of-the-art unified multimodal models and several task-specific understanding models on visual understanding benchmarks. It is also capable of generating high-quality images. The resolution of the images is $384 \times 384$. 
}
\vspace{-2pt}
\label{fig:teaser}
\end{figure}

Large language models (LLMs) have demonstrated remarkable capabilities in learning diverse knowledge and generalizing to new scenarios~\citep{touvron2023llama, mann2020language, achiam2023gpt, bubeck2023sparks, 2024DeepSeekLLM}. Leveraging these capabilities, researchers have developed sophisticated models specialized in image comprehension~\citep{2024LLaVA, 2024LLaVA1.5, 2024LLaVAOV, alayrac2022flamingo, 2023instructblip, li2023blip} and text-to-image generation~\citep{2022Imagen, 2022DALLE2, 2023SDXL, 2024SD3}.

The field has recently shifted toward creating unified systems capable of handling both tasks simultaneously. One prominent direction involves utilizing pre-trained text-to-image models for high-quality generation while training LLMs to generate conditions for these models~\citep{2023Seed, 2023SeedLLaMa, 2024Emu, 2024SeedX, 2024dreamllm}. However, this approach introduces architectural complexity and potentially constrains the model's capabilities through maintaining separate LLM and generative components. Alternative approaches~\citep{2024Chameleon, 2024VILAU, 2024Showo, 2024Transfusion, 2024Janus} propose training a single LLM for both tasks, typically incorporating either diffusion models~\citep{2020DDPM, 2021ScoreModel} or vector-quantized autoregressive models~\citep{esser2021taming, 2024llamagen}. 

Our approach builds upon recent breakthroughs in rectified flow models~\citep{2022RF, 2022FlowMatching, albergo2023building, 2024SD3, liu2023instaflow}, which provide a simple framework for generative modeling while delivering exceptional empirical performance~\citep{2024SD3, le2024voicebox, jin2024pyramidal}.
Building on these advances, we propose \textbf{\ours}, a powerful unified multimodal model that seamlessly integrates rectified flow with LLM architecture.
Following a minimalist design principle, our architecture requires only a lightweight encoder and decoder to adapt the LLM for rectified flow operations.
To optimize \ours's performance, we implement two key strategies: First, we maintain separate vision encoders for understanding and generation tasks, preventing task interference and thus enhancing comprehension capabilities. Second, we align the intermediate representations between generation and understanding modules during training, strengthening semantic coherence in the generation process.

\ours~shows state-of-the-art performances in both multimodal comprehension and text-to-image generation compared to existing unified approaches, and even outperforms several specialized methods. Specifically, on text-to-image generation benchmarks, MJHQ FID-$30$k \cite{2024PG2.5}, GenEval \cite{2024Geneval} and DPG-Bench \cite{2024DPGBench}, \ours~achieves scores of $9.51$, $0.63$ and $80.09\%$, surpassing established text-to-image models including SDv1.5~\citep{2022LDM} and SDXL~\citep{2023SDXL}. In multimodal comprehension benchmarks, \ours~attains scores of $74.9$, $70.5$ and $60.3$ on MMBench~\citep{2024MMBench}, SeedBench~\citep{2023SeedBench}, and GQA~\citep{2019GQA}, respectively, exceeding specialized models such as LLaVA-v1.5 \cite{2024LLaVA1.5} and Qwen-VL-Chat \cite{2023Qwen}. Notably, these results are achieved with a compact LLM architecture with only 1.3B parameters.

\section{Related Work}

\paragraph{Visual Generation with Flow-based Generative Models.}
Recent years have witnessed remarkable progress in visual generation through diffusion models~\citep{2020DDPM, 2021ScoreModel}, leading to impressive models like~\citep{2022LDM, 2023SDXL, 2022MMDiffusion, 2022Imagen, 2022DALLE2, 2023ummdiffusion}. Building on these advances, flow-based generative models~\citep{2022RF, 2022FlowMatching, albergo2023building} emerged as a simplified alternative framework. These approaches have recently enabled advanced visual generation models~\citep{2024SD3, jin2024pyramidal} that achieve superior empirical performance with faster sampling. Our work demonstrates that rectified flow~\citep{2022RF, liu2022rectified, liu2023instaflow} can be effectively integrated into LLMs, creating unified models that excel in both understanding and generation tasks.

\paragraph{Unified Models For Understanding and Generation.}
The development of multimodal large language models (MLLMs) has enabled effective integration of text and visual information. Building upon powerful LLMs~\citep{touvron2023llama, touvron2023llama2, 2024DeepSeekLLM}, recent MLLMs~\citep{2024DeepSeekVL, 2024LLaVA, 2024LLaVA1.5, alayrac2022flamingo, 2023instructblip, li2023blip} have demonstrated exceptional multimodal understanding capabilities. Current research increasingly focuses on architectures that can simultaneously handle visual understanding and generation tasks. One approach extends MLLMs with pre-trained diffusion models~\citep{2023Seed, 2023SeedLLaMa, 2024Emu, 2024SeedX, 2024dreamllm, ye2024x}. However, these systems essentially utilize diffusion models as external tools, where the MLLM generates conditions for image generation without possessing direct generative capabilities. This separation often results in suboptimal performance compared to standalone diffusion models~\cite{2023Seed, 2024Emu}. Another line of work~\citep{2024Chameleon, 2024VILAU, 2024Showo, 2024Transfusion, 2024Janus} aim to train a single LLM for both tasks. Many of these methods employ vector-quantization~\citep{esser2021taming, 2024llamagen} to convert images into discrete tokens, enabling unified autoregressive processing~\citep{2024Chameleon, 2024Janus}. While straightforward to implement, these approaches are inherently limited by their image tokenization quality.

Our work focuses on developing unified models that combine autoregressive capabilities with flow/diffusion models, leveraging their proven effectiveness in visual generation. Compared to similar approaches~\citep{2024Showo, 2024Transfusion, 2024Monoformer}, \ours~offers three key advantages: (i) a simple yet effective generation process using rectified flow, (ii) enhanced performance through decoupled vision encoders that resolve inter-task conflicts, and (iii) improved generation quality through representation alignment regularization, enabled by our decoupled encoder design.
\section{\ours}

In this section, we introduce the architecture of \ours~and our training strategies.

\subsection{Background}

\paragraph{Multimodal LLMs.} Given a dataset $\D$ containing discrete token sequences, each of which can be formulated as $x = (x_1, \cdots, x_\ell)$, large language models (LLMs) are trained to model the sequence distribution in an autoregressive manner, 
\begin{equation}
\log {\p}_{\theta_{LLM}} (x) = \sum_{i = 0}^{\ell-1} \log {\p}_{\theta_{LLM}}(x_{i+1} | x_{1}, \dots, x_{i}),
\end{equation}
where $\theta_{LLM}$ denotes the parameters of the LLM and $\ell$ is the sequence length. After being trained on large-scale datasets, LLMs exhibit the ability to generalize across various tasks and follow diverse instructions~\citep{bubeck2023sparks, achiam2023gpt, mann2020language}. To extend these models to handle visual inputs, LLMs are augmented with vision encoders~\citep{2024LLaVA, 2024LLaVA1.5, alayrac2022flamingo}. For instance, LLaVA~\citep{2024LLaVA} integrates an LLM with a pre-trained CLIP~\citep{2021CLIP} image encoder via a projection layer, transforming the extracted image features into a joint embedding space that the LLM can process as word embeddings. By leveraging large-scale multimodal datasets and increasingly powerful LLMs, this architecture has facilitated the development of advanced multimodal models capable of addressing a wide range of vision-language tasks~\citep{2023Qwen, 2024LLaVA1.5, 2024LLaVAOV, 2024DeepSeekVL}.

\paragraph{Rectified Flow.} For a dataset $\D$ consisting of continuous $d$-dimensional data points $x = (x_1, \cdots, x_d)$ drawn from an unknown data distribution $\pi_{1}$, rectified flow~\citep{2022RF, 2022FlowMatching} models the data distribution by 
learning an ordinary differential equation (ODE) defined over time $t \in [0,1]$:
\begin{equation}
    \frac{\d z_t}{\d t} = v_{\theta_{NN}}(z_t, t),~~~~~z_0 \sim \pi_{0},
\end{equation}
where $\theta_{NN}$ represents the parameters of the velocity neural network and $\pi_{0}$ is a simple distribution, typically standard Gaussian noise $\mathcal{N}(0, I)$. 
The network is trained by minimizing the Euclidean distance between the neural velocity and the directions of linear paths connecting random points from $\pi_0$ and $\pi_1$,
\begin{equation}
    \min_{\theta} {\E}_{t\sim {\p}(t), z_0 \sim \pi_0, x\sim \pi_1} \left [ \left|\left| v_{\theta_{NN}}(z_t, t) - (x - z_0) \right|\right|^2 \right ],~~\text{where}~~z_t = t x + (1-t) z_0.
\end{equation}
Here, $\p(t)$ is a distribution over time $t \in [0,1]$. When the network has sufficient capacity and the objective is perfectly minimized, the optimal velocity field $v_{\theta^*_{NN}}$ maps the elementary distribution $\pi_0$ to the true data distribution $\pi_1$. More precisely, the distribution of $z_1 = \int_{0}^1 v_{\theta^*_{NN}}(z_t, t) {\d} t$, with $z_0 \sim \pi_0$, follows $\pi_1$. Despite its conceptual simplicity, rectified flow has shown superior performance in various generative modeling tasks, including text-to-image generation~\citep{2024SD3}, audio generation~\citep{kim2024p} and biological structure generation~\citep{jingalphafold}.

\begin{figure}
    \centering
    \includegraphics[width=\linewidth]{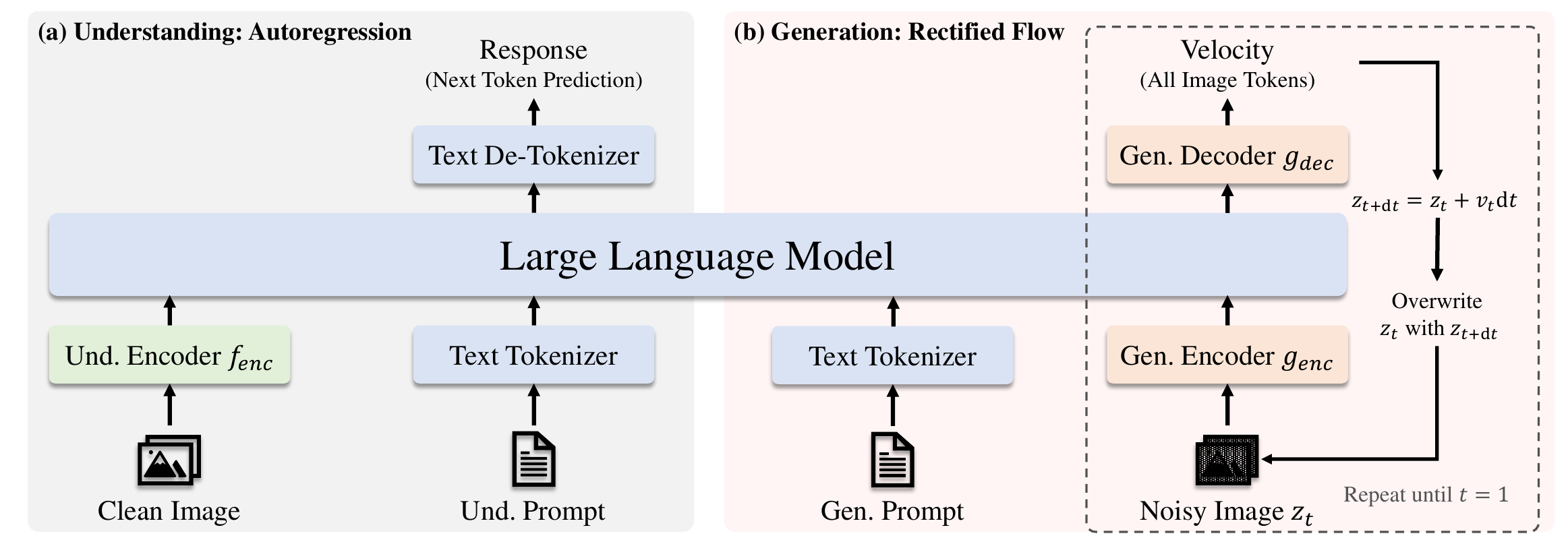}
    \caption{
    \textbf{Architecture of the proposed \ours.} For visual understanding, the LLM performs autoregressive next-token prediction to generate responses. For image generation, the LLM employs images with rectified flow. Starting from Gaussian noise at $t=0$, the LLM iteratively updates $z_t$ by predicting velocity vectors until reaching $t=1$. We omit the VAE encoder, the skip connection leveraged in generation and the linear layer after $f_{enc}$ for simplicity.
    }
    \label{fig:method-arch}
\end{figure}

\subsection{A Unified Framework for Multimodal Understanding and Generation}

\ours~presents a unified framework designed to address both vision understanding and image generation tasks. Next we outline how \ours~handles these two tasks within a single LLM architecture.

\paragraph{Multimodal Understanding.} In multimodal understanding tasks, the LLM processes an input sequence consisting of interleaved text and image data. The text is tokenized into discrete tokens, each of which is transformed into an embedding of dimension $D_{emb}$. For the images, an image encoder $f_{enc}$ encodes each image $x_{im}$ into a feature map of shape $H_{im} \times W_{im} \times D_{enc}$. 
This feature map is flattened and projected through a linear transformation layer into a sequence of embeddings with shape $H_{im} W_{im} \times D_{emb}$. 
$H_{im}$ and $W_{im}$ are determined by the image encoder. The text and image embeddings are concatenated to form the input sequence to the LLM, which then autoregressively predicts the next tokens based on the input sequence of embeddings. According to common practice~\citep{2024Janus, 2024Chameleon, 2024Showo}, we add special token \texttt{|BOI|} before the image and \texttt{|EOI|} after the image to help the model locate the image embeddings in the sequence.

\paragraph{Image Generation.} For image generation, our LLM takes a text sequence $x^{con}$ as condition and generates a corresponding image using rectified flow. 
To improve computational efficiency, generation occurs in the latent space using a pre-trained SDXL-VAE~\citep{2023SDXL}.  

The generation process begins by sampling Gaussian noise $z_0$ of shape $H_{latent} \times W_{latent} \times D_{latent}$ in the latent space, which is then processed by a generation encoder $g_{enc}$ into a sequence of embeddings $H_{gen} W_{gen} \times D_{emb}$. This sequence is concatenated with a time embedding representing the current time step $t$ ($t=0$ at the beginning), resulting in a sequence of length $H_{gen} W_{gen}+1$. 
Unlike previous approaches that employ various attention masking strategies~\citep{2024Showo, 2024Transfusion}, we found that causal attention suffices, as our preliminary experiments showed no performance benefits from alternative masking schemes.
The LLM's output corresponding to $z_0$ is transformed back into the latent space by a generation decoder $g_{dec}$, producing a velocity vector of shape $H_{latent} \times W_{latent} \times D_{latent}$. The state is updated by a standard Euler solver,
\begin{equation}
    z_{t + \d t} = z_t + v(z_t, t) \d t, 
\end{equation}
where $\d t$ is a user-defined step size.  We replace $z_0$ with $z_{\d t}$ on the input and iterate the process until we get $z_1$, which is then decoded into the final image by the VAE decoder. To enhance generation quality, we employ classifier-free guidance (CFG) when computing the velocity:
\begin{equation}
v(z_t, t) = w v(z_t, t~|~x^{con}) + (1-w) v(z_t, t~|~\varnothing),
\end{equation}
where $v(z_t, t~|~\varnothing)$ denotes the velocity inferred without text conditioning and $w \geq 1$ controls the magnitute of CFG. Empirically, increasing $w$ yields higher semantic alignment~\citep{2022LDM, liu2023instaflow, 2023SDXL, 2024SD3}. Analogous to multimodal understanding, we prepend the special token \texttt{|BOI|} to indicate the start of image generation in the sequence.

\paragraph{Decoupling Encoders for the Two Tasks.}

Previous approaches that unify autoregressive generation and diffusion models within a joint LLM training framework~\citep{2024Transfusion, 2024Showo} employ identical encoders ($f_{enc}$ and $g_{enc}$) for both understanding and generation tasks. For instance, \citet{2024Transfusion} performs both tasks in the same VAE latent space using a shared U-Net or linear encoder, while \citet{2024Showo} leverages MAGVIT-v2~\citep{yulanguage} to encode image patches into discrete tokens for both tasks. 

However, recent work on unified autoregressive models has shown this shared encoder design to be suboptimal~\citep{2024Janus}, particularly in models that generate images through autoregression on vector-quantized tokens. Drawing from these insights, \ours~adopts a decoupled encoder design. Specifically, we employ a pre-trained SigLIP-Large-Patch/16~\citep{2023SigLIP} model as $f_{enc}$ to extract semantic continuous features for multimodal understanding, while using separate ConvNeXt blocks~\citep{2023Convnext} initialized from scratch as $g_{enc}$ and $g_{dec}$ for generation, chosen for its effectiveness. Following established practices~\citep{2023UViT, 2024HDiT, 2024Vermeer}, we incorporate a long skip connection between $g_{enc}$ and $g_{dec}$. Our controlled experiments in Sec.~\ref{sec exp, subsec ablate} demonstrate that this decoupled encoder design significantly improves the performance of our unified model. The complete architecture of \ours~is illustrated in Fig.~\ref{fig:method-arch}.

\subsection{Training Schemes}

\begin{figure}
    \centering
    \includegraphics[width=\linewidth]{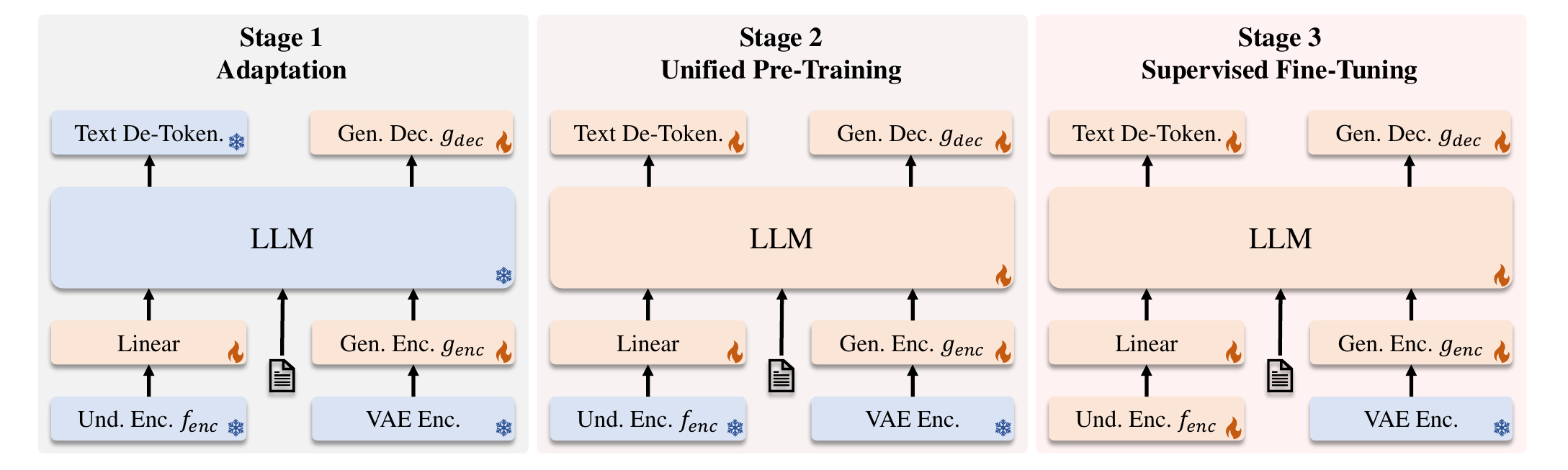}
    \caption{\textbf{Three training stages of \ours.} 
    The trainable modules are marked with flame and the frozen modules are marked with snowflakes.}
    \label{fig:method-phase}
\end{figure}

As illustrated in Fig.~\ref{fig:method-phase}, we train our model in three sequential stages, detailed below.

\paragraph{Stage 1: Adaptation of Randomly Initialized Components.} In the first stage, we focus on training only the randomly initialized components: the linear layers, generation encoder, and generation decoder. This stage serves to adapt these new modules to work effectively with the pre-trained LLM and SigLIP encoder, essentially functioning as an initialization phase for the newly introduced components.

\paragraph{Stage 2: Unified Pre-Training.} Following the adaptation stage, we train the entire model except for the visual encoder, consistent with previous approaches~\cite{2024LLaVA, 2024DeepSeekVL}. The training incorporates three data types: multimodal understanding, image generation, and text-only data. We initially allocate a higher proportion of multimodal understanding data to establish the model's understanding capabilities. Subsequently, we increase the ratio of image generation data to accommodate the convergence requirements of diffusion-based models~\cite{2021ADM, 2023DiT}.

\paragraph{Stage 3: Supervised Fine-Tuning (SFT).} In the final stage, we fine-tune the pre-trained model using instruction tuning data, which comprises dialogues, task-specific conversations, and high-quality text-conditioned image generation examples. 
During this stage, we also unfreeze the SigLIP encoder parameters~\citep{2024DeepSeekVL, tong2024cambrian, 2024Janus}. This fine-tuning process enables the model to effectively respond to user instructions for both multimodal understanding and image generation tasks.

\subsection{Training Objective}
\label{sec method, subsec train}

Training \ours~involves two types of data, multimodal understanding data and image generation data. 
Both types of data contain two parts: ``condition'' and ``response''. ``Condition'' refers to the prompting of the tasks (\eg text prompts in the task of generation and images in the task of understanding) while ``response'' refers to the corresponding responses of the two tasks. The data can be formatted as $x = (x^{con}, x^{res})$, where the superscript $con$ denotes ``condition'' and $res$ denotes ``response''.
We denote the length of the whole sequence $x$ as $\ell$, the length of $x^{con}$ as $\ell_{con}$ and the length of $x^{res}$ as $\ell_{res}$. 
We use $\theta$ to represent the collection of all the trainable parameters in \ours, including the LLM, $f_{enc}$, $g_{enc}$, $g_{dec}$ and the linear transformation layers.

\paragraph{Autoregression Objective.} 
For mutimodal understanding tasks, $x^{res}$ contains only text tokens. \ours~is trained using the maximum likelihood principle,
\begin{equation}
    \mathcal{L}_{AR}(\theta) = - \E_{x \sim \D_{und}} \left[ \sum_{i = \ell_{con}}^{\ell-1} \log {\p}_\theta \left(x_{i+1} | x_{1}, \dots, x_{i} \right) \right ],
\end{equation}
where the expectation is taken over all $(x^{con}, x^{res})$ pairs in our multimodal understanding dataset $\mathcal{D}_{und}$, computing loss only over tokens in $x^{res}$.

\paragraph{Rectified Flow Objective.} For image generation tasks, $x^{con}$ consists of text tokens and $x^{res}$ is the corresponding image. \ours~is trained with the rectified flow objective,
\begin{equation}
    \mathcal{L}_{RF}(\theta) = {\E}_{x \sim \D_{gen}, t\sim {\p}(t), z_0 \sim \mathcal{N}(0, I)} \left [ \left|\left| v_\theta(z_t, t~|~x^{con}) - (x^{res} - z_0) \right|\right|^2 \right ],
\end{equation}
where $z_t = t x^{res} + (1-t) z_0$. 
Following Stable Diffusion 3~\cite{2024SD3}, we set the time distribution $\p(t)$ to the logit-normal distribution. To enable CFG inference, we randomly drop 10\% of the text prompts in training. 

\paragraph{Representation Alignment Regularization.}
Recent work~\cite{2024REPA} has shown that aligning intermediate representations between diffusion transformers and semantic vision encoders enhances diffusion model generalization.
Our decoupled vision encoder design enables efficient implementation of this alignment as a regularization term. 
Specifically, for generation tasks, we align features from the understanding encoder $f_{enc}$ with the LLM's intermediate features,
\begin{equation}
    \mathcal{L}_{REPA}(\theta, \phi) = - \E_{x \sim \D_{gen}} \left [ \text{sim} \left (\text{\texttt{stop\_grad}}(f_{enc}(x^{res})), h_\phi(q_\theta(z_t)) \right) \right ],
\end{equation}
where $q_\theta(z_t)$ denotes an intermediate LLM representation given input $z_t$, and $h_\phi$ is a small trainable MLP that projects $q_\theta(z_t)$ to dimension $D_{enc}$. The function $\text{sim}(\cdot, \cdot)$ computes the mean of element-wise cosine similarity between embeddings. Before computing the loss, we reshape $h_\phi(q_\theta(z_t))$ to $H_{gen} \times W_{gen} \times D_{enc}$. To simplify the implementation, we intentionally adjust the configuration of $g_{enc}$ and $g_{dec}$ to ensure $H_{gen}=H_{im}$ and $W_{gen}=W_{im}$.
The gradient of $\mathcal{L}_{REPA}$ is not back-propagated through the understanding encoder. 
This alignment loss helps the LLM's internal feature space (given noisy input $z_t$) align with the understanding encoder's semantic feature space, thereby improving generation quality when producing images from new random noise and text conditions during inference.

\paragraph{Summary.} All three objectives are applied across all training stages. Multimodal understanding tasks use $\mathcal{L}_{AR}$, while image generation tasks employ the combined loss $\mathcal{L}_{RF} + \mathcal{L}_{REPA}$. Detailed experimental settings are provided in Sec.~\ref{sec exp, subsec imple}.

\begin{table}[t]
    \centering
    \small
    \caption{\textbf{Hyper-parameters of the proposed \ours}. Data ratio denotes the proportion of multimodal understanding data, image generation data and text-only data. In the initial $10,000$ steps of Stage 2, we apply a data ratio of $30:50:20$ to boost the understanding ability.}
    \begin{tabular}{l|cccc}
        \toprule
         & \textbf{Stage 1} & \textbf{Stage 2} & \textbf{Stage 3} \\
        \midrule
        Learning Rate & $1.0\times10^{-4}$ & $1\times10^{-4}$ & $2.0\times10^{-5}$  \\
        LR Scheduler  & Constant &  Constant & Constant \\
        Weight Decay  & $0.0$ & $0.0$ & $0.0$ \\
        Gradient Clip & $1.0$ & $1.0$ & $1.0$ \\
        Optimizer     & \multicolumn{3}{c}{AdamW ($\beta_1=0.9, \beta_2=0.95$)} \\
        Warm-up Steps    & $2,000$ & $2,000$ & $1,000$ \\
        Training Steps   & $10,000$ & $390,000$ & $26,000$ \\
        Batch Size       & $512$ & $512$ & $256$ \\
        Data Ratio       & $50:50:0$ & $14:80:6$ & $21:70:9$ \\
        \bottomrule
    \end{tabular}
    \label{tab:exp-phase}
\end{table}

\section{Experiments}

We conduct extensive experiments to evaluate the capabilities of \ours~in both multimodal understanding and generation tasks. First, we describe our experimental setup and implementation details. Then, we present results on standard benchmarks for multimodal understanding and image generation. Finally, we perform ablation studies to validate our key design choices.

\subsection{Experiment Setup and Implementation Details}
\label{sec exp, subsec imple}

Our framework builds upon an enhanced version\footnote{This version, trained on an expanded text corpus compared to the one in Janus~\cite{2024Janus}, has been demonstrated to possess better performance on multiple-choice benchmarks (e.g., MMBench~\cite{2024MMBench} and SEED Bench~\cite{2023SeedBench}). Our preliminary experiments suggest that it has minimal impact on the quality of visual generation.} of DeepSeek-LLM (1.3B)~\cite{2024DeepSeekLLM, 2024DeepSeekVL}. The LLM consists of 24 transformer blocks and supports a sequence length of $4,096$. 
In our model, both understanding and generation exploits images of resolution 384.

For multimodal understanding, we leverage SigLIP-Large-Patch/16~\cite{2023SigLIP} as $f_{enc}$. 
For image generation, we utilize the pre-trained SDXL-VAE~\cite{2023SDXL} for its latent space. 
The generation encoder $g_{enc}$ comprises a $2\times2$ patchify layer followed by two ConvNeXt~\cite{2023Convnext} blocks and a linear layer. The generation decoder $g_{dec}$ combines two ConvNeXt blocks, a pixel-shuffle layer to upsample the feature map, and a linear layer. Our SigLIP encoder contains $\sim 300$M parameters. $g_{enc}$ and $g_{dec}$ are light-weight modules, containing $\sim 70$M parameters in total. Table~\ref{tab:exp-phase} details the hyperparameters for each training stage. In the alignment regularization, we use the LLM features after the 6th block as $q_\theta(z_t)$ and a three-layer MLP as $h_\varphi$. We employ an exponential moving average (EMA) with a ratio of 0.99 to ensure training stability.

For data preprocessing, we deal with understanding and generation data differently. For understanding tasks, we maintain all image information by resizing the long side to the target size and padding the image to squares. For generation tasks, we resize the short side to the target size and apply random square cropping to avoid padding artifacts. During training, multiple sequences are packed to form a single sequence of length $4,096$ for training efficiency. Our implementation is based on the HAI-LLM platform~\citep{2023HAI-LLM} using PyTorch~\citep{2023PyTorch}. Training was conducted on NVIDIA A100 GPUs, with each model requiring $\sim1,600$ A100 GPU days.

\subsection{Training Data Settings}

We follow Janus~\cite{2024Janus} to construct the training data. The data configuration for each training stage is listed below.

\paragraph{Data for Stage 1 and Stage 2.} The first two stages of our framework uses three types of data: multimodal understanding data, image generation data and text-only data.
\begin{enumerate}
    \item \textbf{Multimodal Understanding Data.} This type of data contains several sub-categories: (a) Image caption data. We incorporate caption datasets from \cite{2023DetailedCaption, 2023SAM, 2024ArxivQA, 2024DenseFusion, 2024MMSci, 2024PixelProse} and 
    generate additional captions for images from \cite{2022LA, 2020OpenImgsv4} using open-source multimodal understanding models. The names of the datasets are provided in the supplementary materials.
    The data follows template formats,~\eg``\texttt{<image>}\texttt{Generate the caption of this picture.} \texttt{<caption>}''. 
    (b) Charts and tables. We directly adopt the chart and table data from the training data of DeepSeek-VL \cite{2024DeepSeekVL}. 
    (c) Task data. 
    ShareGPT4V \cite{2023Sharegpt4v} data is utilized to facilitate basic question-answering capabilities during pre-training, structured as ``\texttt{<image>}\texttt{<question>}\texttt{<answer>}''.
    (d) Interleaved text-image data. This sub-category is sourced from \cite{2021wit, 2018wikihow}.
    \item \textbf{Image Generation Data.} 
    Our image generation dataset combines high-quality images from \cite{2022LA, 2023SAM, 2020OpenImgsv4, 2024Megalith, 2024Yfcc, 2024JDB, 2024DALLE3-Img, 2024PixelProse} and $2$ million in-house data.
    We enhance them with machine-generated captions using multimodal understanding models.
    We filter the images in \cite{2022LA, 2024PixelProse} with aspect ratios and aesthetic scores, retaining approximately $20\%$ of the original datasets. 
    $25\%$ of the data contains single-sentence captions. These kind of data assist the model to be able to process short prompts. All the data points are formatted as ``\texttt{<prompt>}\texttt{<image>}''.
    \item \textbf{Text-Only Data.} We directly use the text corpus of DeepSeek-LLM \cite{2024DeepSeekLLM}.
\end{enumerate}

\noindent \textbf{Data for Stage 3.} The SFT stage also uses three types of data:
\begin{enumerate}
    \item \textbf{Multimodal Instruction Data.} We leverage the instruction tuning datasets from \cite{2017VQAv2, 2022screenqa, 2019GQA, 2024LLaVAOV, 2021iconqa, 2019kvqa}.
    \item \textbf{Image Generation Data.} 
    We reformat the high-quality text-image pairs from~\cite{2022LA, 2024JDB, 2024PixelProse} into an instruction format: ``\texttt{User:}\texttt{<user prompt>}\texttt{\textbackslash n\textbackslash n} \texttt{Assistant:<image>}''.
    \item \textbf{Text-Only Data.} We directly incorporate the text-only data from \cite{2024LLaVAOV}.
\end{enumerate}

\begin{table}[t]
    \centering
    \setlength{\tabcolsep}{4pt}
    \renewcommand{\arraystretch}{1.2}
    \scriptsize
    \caption{\textbf{Performances on GenEval benchmark.} ``Gen.'' denotes ``generation'' and ``Unified'' denotes unified understanding and generation models. Models using external pre-trained generative models are signed with $^\dagger$. 
    }
    \begin{tabular}{llcccccccc}
        \toprule
        \textbf{Type} & \textbf{Method} & \textbf{Params} & \textbf{Single Obj.} & \textbf{Two Obj.} & \textbf{Count.} & \textbf{Colors} & \textbf{Pos.} & \textbf{Color Attri.} & \textbf{Overall$\uparrow$} \\
        \midrule
        \multirow{10}{*}{\textit{Gen. Only}} 
        & LlamaGen~\cite{2024llamagen} & $0.8$B & $0.71$ & $0.34$ & $0.21$ & $0.58$ & $0.07$ & $0.04$ & $0.32$ \\
        & LDM~\cite{2022LDM} & $1.4$B & $0.92$ & $0.29$ & $0.23$ & $0.70$ & $0.02$ & $0.05$ & $0.37$ \\
        & SDv$1.5$~\cite{2022LDM} & $0.9$B & $0.97$ & $0.38$ & $0.35$ & $0.76$ & $0.04$ & $0.06$ & $0.43$ \\
        & PixArt-$\alpha$~\cite{2023Pixelartalpha} & $0.6$B & $0.98$ & $0.50$ & $0.44$ & $0.80$ & $0.08$ & $0.07$ & $0.48$ \\
        & SDv$2.1$~\cite{2022LDM} & $0.9$B & $0.98$ & $0.51$ & $0.44$ & $0.85$ & $0.07$ & $0.17$ & $0.50$ \\
        & DALL-E $2$~\cite{2022DALLE2} & $6.5$B & $0.94$ & $0.66$ & $0.49$ & $0.77$ & $0.10$ & $0.19$ & $0.52$ \\
        & Emu$3$-Gen ~\cite{2024emu3} & $8$B & $0.98$ & $0.71$ & $0.34$ & $0.81$ & $0.17$ & $0.21$ & $0.54$ \\
        & SDXL~\cite{2023SDXL} & $2.6$B & $0.98$ & $0.74$ & $0.39$ & $0.85$ & $0.15$ & $0.23$ & $0.55$ \\
        & IF-XL~\cite{2023IF} & $4.3$B & $0.97$ & $0.74$ & $0.66$ & $0.81$ & $0.13$ & $0.35$ & $0.61$ \\
        & DALL-E $3$~\cite{2023dalle3} & - & $0.96$ & $0.87$ & $0.47$ & $0.83$ & $0.43$ & $0.45$ & $0.67$ \\
        \midrule
        \multirow{7}{*}{\textit{Unified}}
        & Chameleon~\cite{2024Chameleon} & $34$B & - & - & - & - & - & - & $0.39$ \\
        & LWM~\cite{2024LWM} & $7$B & $0.93$ & $0.41$ & $0.46$ & $0.79$ & $0.09$ & $0.15$ & $0.47$ \\
        & SEED-X$^\dagger$~\cite{2024SeedX} & $17$B & $0.97$ & $0.58$ & $0.26$ & $0.80$ & $0.19$ & $0.14$ & $0.49$ \\
        & Show-o~\cite{2024Showo} & $1.3$B & $0.95$ & $0.52$ & $0.49$ & $0.82$ & $0.11$ & $0.28$ & $0.53$ \\
        & Janus \cite{2024Janus} & $1.3$B & $0.97$ & $0.68$ & $0.30$ & $0.84$ & $0.46$ & $0.42$ & $0.61$ \\
        & Transfusion~\cite{2024Transfusion} & $7.3$B & - & - & - & - & - & - & 0.63 \\
        & \textbf{\ours~(Ours)} & 1.3B & 0.97 & 0.59 & 0.45 & 0.83 & 0.53 & 0.42 & 0.63 \\
        \bottomrule
    \end{tabular}
    \label{tab:exp-geneval}
\end{table}

\begin{table}[t]
    \centering
    \renewcommand{\arraystretch}{1.2}
    \scriptsize
    \caption{\textbf{Performances on DPG-Bench.} The methods in this table are all generation-specific models except our method.}
    \begin{tabular}{lcccccc}
        \toprule
        \textbf{Method} & \textbf{Global} & \textbf{Entity} & \textbf{Attribute} & \textbf{Relation} & \textbf{Other} & \textbf{Overall$\uparrow$} \\
        \midrule
        SDv1.5 \cite{2022LDM} & 74.63 & 74.23 & 75.39 & 73.49 & 67.81 & 63.18 \\
        PixArt-$\alpha$ \cite{2023Pixelartalpha} & 74.97 & 79.32 & 78.60 & 82.57 & 76.96 & 71.11 \\
        Lumina-Next \cite{2024lumina} & 82.82 & 88.65 & 86.44 & 80.53 & 81.82 & 74.63 \\
        SDXL \cite{2023SDXL} & 83.27 & 82.43 & 80.91 & 86.76 & 80.41 & 74.65 \\
        Playground v2.5 \cite{2024PG2.5} & 83.06 & 82.59 & 81.20 & 84.08 & 83.50 & 75.47 \\
        Hunyuan-DiT \cite{2024hunyuandit} & 84.59 & 80.59 & 88.01 & 74.36 & 86.41 & 78.87 \\
        PixArt-$\Sigma$ \cite{2024pixartsigma} & 86.89 & 82.89 & 88.94 & 86.59 & 87.68 & 80.54\\
        Emu3-Gen \cite{2024emu3} & 85.21 & 86.68 & 86.84 & 90.22 & 83.15 & 80.60 \\
        \textbf{\ours~(Ours)} & 87.03 & 87.31 & 87.39 & 89.79 & 88.10 & 80.09 \\
        \bottomrule
    \end{tabular}
    \label{tab:exp-dpg}
\end{table}

\subsection{Evaluation Settings}

\paragraph{Image Generation.} 
We evaluate the generated images using both visual quality and semantic accuracy metrics. For visual quality assessment, we employ the Fréchet Inception Distance \cite{2017TTUR&FID} (FID) metric and compute FID between 30,000 generated images and their corresponding reference images from the MJHQ dataset~\cite{2024PG2.5}. The FID computation follows the implementation from GigaGAN~\cite{2023gigagan}. To evaluate semantic accuracy, we utilize two specialized frameworks: GenEval \cite{2024Geneval} and DPG-Bench \cite{2024DPGBench}. These frameworks are designed to assess whether the generated images accurately contain the objects and relationships specified in the input prompts, providing a broad evaluation of the generation capabilities.

\paragraph{Multimodal Understanding.} We evaluate \ours's multimodal understanding abilities across a diverse set of vision-language benchmarks for general understanding capabilities, including POPE \cite{2023POPE}, MME \cite{2023MME}, MMBench \cite{2024MMBench}, SEEDBench \cite{2023SeedBench}, VQAv2 \cite{2017VQAv2}, GQA \cite{2019GQA}, MM-Vet \cite{2024mmvet}, MMMU \cite{2024mmmu}, ChartQA\cite{2022ChartQA} and TextVQA\cite{2019textvqa}

\subsection{Quantitative Results}

\begin{wraptable}{r}{0.34\textwidth}  
    \centering
    \renewcommand{\arraystretch}{1.2}
    \scriptsize
    \vspace{-2mm}
    \caption{\textbf{Results of MJHQ FID-30k.} 
    The models which have similar scales to our model are marked with blue background.
    \ours~achieves the best FID among 1.3B models. 
    }
    \begin{tabular}{lcc}
        \toprule
        \textbf{Method} & \textbf{Params} & \textbf{FID$\downarrow$}
        \\
        \midrule
        LWM \cite{2024LWM} & 7B & 17.77 \\
        VILA-U 256 \cite{2024VILAU} & 7B & 12.81 
        \\
        VILA-U 384 \cite{2024VILAU} & 7B & 7.69 
        \\
        \cellcolor{blue!3}Show-o \cite{2024Showo} & \cellcolor{blue!3}1.3B & \cellcolor{blue!3}15.18 
        \\
        \cellcolor{blue!3}Janus \cite{2024Janus} & \cellcolor{blue!3}1.3B & \cellcolor{blue!3}10.10 
        \\
        \cellcolor{blue!3}\textbf{\ours~(Ours)} & \cellcolor{blue!3}1.3B & \cellcolor{blue!3}9.51
        \\
        \bottomrule
    \end{tabular}
    \vspace{-1mm}
    \label{tab:exp-mjhq}
\end{wraptable}
\textbf{Image Generation Performances.} We report the performances on GenEval, DPG-Bench and MJHQ FID-30k. In Tab.~\ref{tab:exp-geneval}, we give comparisons on GenEval including the scores of all the sub-tasks and the overall score. \ours~achieves an overall score of 0.63, surpassing the previous unified framework and several generation specific models including SDXL \cite{2023SDXL} and DALL-E 2 \cite{2022DALLE2}. In Tab.~\ref{tab:exp-dpg}, We show results on DPG-Bench and the corresponding comparisons. It is noted that all the methods in Tab.~\ref{tab:exp-dpg} are generation-specific models except our model. The results on GenEval and DPG-Bench demonstrate the ability of instruction following of our model.
We give the comparisons on MJHQ FID-30k in Tab.~\ref{tab:exp-mjhq}. The images which are sampled to calculate FID are generated with a CFG factor $w=2$ and a number of sampling steps $30$. We sweep the CFG factor and the sampling steps and provide the results in the appendix.
Our method achieves the best performance among all the models with 1.3B LLM. The results prove that the rectified flow is able to improve the quality of generated images over autoregressive models such as Janus \cite{2024Janus}.

\noindent \textbf{Multimodal Understanding Performances.} We show comparisons of our method and other methods including understanding-specific models and unified understanding and generation models in Tab.~\ref{tab:exp-com}. Our model reaches the best performances among all the models with similar number of parameters and even surpasses multiple understanding-specific methods with larger scales. Our results demonstrate that our method harmonizes autoregressive LLM and rectified flow, achieving satisfying performance in both understanding and generation.

\begin{table}[t]
    \centering
    \setlength{\tabcolsep}{3.5pt}
    \tiny
    \caption{\textbf{Comparison with other methods on multimodal understanding benchmarks}. ``Und.'' denotes ``understanding'' and ``Unified'' denotes unified understanding and generation models. The models employing external pre-trained generative models are marked with $^\dagger$. The models with LLMs which have similar number of parameters to us are marked with blue background under the line of dashes.
    }
    \vspace{-2mm}
    \label{tab:exp-com}
    \begin{tabular}{llccccccccccc}
        \toprule
        \textbf{Type} & \textbf{Model} & \textbf{LLM Param} & \textbf{POPE} & \textbf{MME-P} & \textbf{MMB\textsubscript{dev}} & \textbf{SEED} & \textbf{VQAv2\textsubscript{test}} & \textbf{GQA} & \textbf{MMMU} & \textbf{MM-Vet} & \textbf{ChartQA} & \textbf{TextVQA}
        \\
        \midrule
        \multirow{15}{*}{\textit{Und. Only}} 
        & MobileVLM~\cite{2023mobilevlm} & $2.7$B & $84.9$ & $1288.9$ & $59.6$ & - & - & $59.0$ & - & - & - & $47.5$
        \\
        & MobileVLM-V2~\cite{2024mobilevlmv2} & $2.7$B & $84.7$ & $1440.5$ & $63.2$ & - & - & $61.1$ & - & - & - & $57.5$
        \\
        & LLaVA-Phi~\cite{2024llavaphi} & $2.7$B & $85.0$ & $1335.1$ & $59.8$ & - & $71.4$ & - & - & $28.9$ & - & $48.6$
        \\
        & LLaVA~\cite{2024LLaVA} & $7$B & $76.3$ & $809.6$ & $38.7$ & $33.5$ & - & - & - & $25.5$ & - & -
        \\
        & LLaVA-v1.5~\cite{2024LLaVA1.5}& $7$B & $85.9$ & $1510.7$ & $64.3$ & $58.6$ & $78.5$ & $62.0$ & $35.4$ & $31.1$ & - & $58.2$
        \\
        & InstructBLIP~\cite{2023instructblip} & $7$B & - & - & $36.0$ & $53.4$ & - & $49.2$ & - & $26.2$ & - & $50.1$
        \\
        & Qwen-VL-Chat~\cite{2023Qwen} & $7$B & - & $1487.5$ & $60.6$ & $58.2$ & $78.2$ & $57.5$ & - & - & $66.3$ & $61.5$
        \\
        & LLaVA-NeXT~\cite{2024llavanext} & $7$B & - & $1519.3$ & - & - & - & - & $35.1$ & - & $54.8$ & -
        \\
        & Qwen2-VL~\cite{2024qwen2vl} & $7$B & - & - & - & - & - & - & $54.1$ & $62.0$ & $83.0$ & $84.3$
        \\
        & IDEFICS-$9$B~\cite{2023idefics} & $8$B & - & - & $48.2$ & - & $50.9$ & $38.4$ & - & - & - & $25.9$
        \\
        & Emu$3$-Chat~\cite{2024emu3} & $8$B & $85.2$ & - & $58.5$ & $68.2$ & $75.1$ & $60.3$ & $31.6$ & - & $68.6$ & $64.7$
        \\
        & InstructBLIP~\cite{2023instructblip} & $13$B & $78.9$ & $1212.8$ & - & - & - & $49.5$ & - & $25.6$ & - & $50.7$
        \\
        \cdashline{2-13}
        \\[-2.5ex]
        & 
        \cellcolor{blue!3}LLaVA-v1.5-Phi-1.5~\cite{2024Showo} & \cellcolor{blue!3}$1.3$B & \cellcolor{blue!3}$84.1$ & \cellcolor{blue!3}$1128.0$ & \cellcolor{blue!3}- & \cellcolor{blue!3}- & \cellcolor{blue!3}$75.3$ & \cellcolor{blue!3}$56.5$ & \cellcolor{blue!3}$30.7$ & \cellcolor{blue!3}- & \cellcolor{blue!3}- & \cellcolor{blue!3}- 
        \\
        & \cellcolor{blue!3}MobileVLM~\cite{2023mobilevlm} & \cellcolor{blue!3}$1.4$B & \cellcolor{blue!3}$84.5$ & \cellcolor{blue!3}$1196.2$ & \cellcolor{blue!3}$53.2$ & \cellcolor{blue!3}- & \cellcolor{blue!3}- & \cellcolor{blue!3}$56.1$ & \cellcolor{blue!3}- & \cellcolor{blue!3}- & \cellcolor{blue!3}- & \cellcolor{blue!3}$41.5$
        \\
        & \cellcolor{blue!3}MobileVLM-V2~\cite{2024mobilevlmv2} & \cellcolor{blue!3}$1.4$B & \cellcolor{blue!3}$84.3$ & \cellcolor{blue!3}$1302.8$ & \cellcolor{blue!3}$57.7$ & \cellcolor{blue!3}- & \cellcolor{blue!3}- & \cellcolor{blue!3}$59.3$ & \cellcolor{blue!3}- & \cellcolor{blue!3}- & \cellcolor{blue!3}- & \cellcolor{blue!3}$52.1$
        \\
        \midrule
        \multirow{11}{*}{\textit{Unified}}
        & Gemini-Nano-1~\cite{2023gemini} & $1.8$B & - & - & - & - & $62.7$ & - & $26.3$ & - & $53.6$ & $62.5$\\
        & LWM~\cite{2024LWM} & $7$B & $75.2$ & - & - & - & $55.8$ & $44.8$ & - & $9.6$ & - & -
        \\
        & VILA-U~\cite{2024VILAU} & $7$B & $85.8$ & $1401.8$ & - & $59.0$ & $79.4$ & $60.8$ & - & $33.5$ & - & $60.8$
        \\
        & Chameleon~\cite{2024Chameleon} & $7$B & - & - & - & - & - & - & $22.4$ & $8.3$ & - & -
        \\
        & DreamLLM$^\dagger$~\cite{2024dreamllm} & $7$B & - & - & - & - & $72.9$ & - & - & $36.6$ & - & $41.8$
        \\
        & LaVIT$^\dagger$~\cite{2024lavit} & $7$B & - & - & - & - & $66.0$ & $46.8$ & - & - & - & -
        \\
        & Emu$^\dagger$~\cite{2024Emu} & $13$B & - & - & - & - & $52.0$ & - & - & - & - & -
        \\
        & NExT-GPT$^\dagger$~\cite{2023nextgpt} & $13$B & - & - & - & - & $66.7$ & - & - & - & - & -
        \\
        \cdashline{2-13}
        \\[-2.5ex]
        & \cellcolor{blue!3}Show-o~\cite{2024Showo} & \cellcolor{blue!3}$1.3$B & \cellcolor{blue!3}$73.8$ & \cellcolor{blue!3}$948.4$ & \cellcolor{blue!3}- & \cellcolor{blue!3}- & \cellcolor{blue!3}$59.3$ & \cellcolor{blue!3}$48.7$ & \cellcolor{blue!3}$25.1$ & \cellcolor{blue!3}- & \cellcolor{blue!3}- & \cellcolor{blue!3}-
        \\
        & \cellcolor{blue!3}Janus \cite{2024Janus} & \cellcolor{blue!3}$1.3$B & \cellcolor{blue!3}$87.0$ & \cellcolor{blue!3}$1338.0$ & \cellcolor{blue!3}$69.4$ & \cellcolor{blue!3}$63.7$ & \cellcolor{blue!3}$77.3$ & \cellcolor{blue!3}$59.1$ & \cellcolor{blue!3}$30.5$ & \cellcolor{blue!3}$34.3$ & \cellcolor{blue!3}- & \cellcolor{blue!3}-
        \\
        & \cellcolor{blue!3}\textbf{\ours~(Ours)} & \cellcolor{blue!3}$1.3$B & \cellcolor{blue!3}$88.0$ & \cellcolor{blue!3}$1333.1$ & \cellcolor{blue!3}$74.9$ & \cellcolor{blue!3}$70.5$ & \cellcolor{blue!3}$79.8$ & \cellcolor{blue!3}$60.3$ & \cellcolor{blue!3}$29.3$ & \cellcolor{blue!3}$30.9$ & \cellcolor{blue!3}$64.6$ & \cellcolor{blue!3}$55.5$
        \\
        \bottomrule
    \end{tabular}
    \vspace{-2mm}
\end{table}

\subsection{Ablation Studies}
\label{sec exp, subsec ablate}

\begin{table}[t]
    \centering
    \setlength{\tabcolsep}{2.3pt}
    \scriptsize
    \caption{\textbf{Ablation studies.} The weights of the modules with $^\dagger$ are frozen during training. ``Exp.'' denotes ``experiment''. ``FID'' in this table is MJHQ FID-10k with CFG factor $w=7.5$ and 30 steps. ``CLIP'' denotes CLIP similarity with the backbone of CLIP-ViT-Large-Patch/14. Exp. F is the final configuration for training \ours.} 
    \begin{tabular}{cccccccccccc}
        \toprule
        \multirow{2}{*}{\textbf{Exp. ID}} & \multicolumn{4}{c}{\textbf{Model Setting}} & \multirow{2}{*}{\textbf{Train. Iter.}} & \multicolumn{5}{c}{\textbf{Evaluation Benchmarks}} \\
        & \textbf{REPA} & \textbf{Und. Modules} & \textbf{Gen. Modules} & \textbf{Type} &  & \textbf{POPE$\uparrow$} & \textbf{VQAv2$_{val}\uparrow$} & \textbf{GQA$\uparrow$} & \textbf{FID$\downarrow$} & \textbf{CLIP $\uparrow$}\\
        \midrule
        A & $\times$ & SigLIP & VAE$^\dagger$+ConvNeXt  & Unified & 50,000 & 82.40 & 69.62 & 54.43 & 19.84 & 24.94 \\
        \midrule
        B & $\checkmark$ & \multicolumn{2}{c}{Shared VAE$^\dagger$+ConvNeXt}  & Unified & 50,000 & 78.13 & 53.94 & 44.04 & 18.05 & 26.38 \\
        C & $\checkmark$ & VAE+ConvNeXt & VAE$^\dagger$+ConvNeXt & Unified & 50,000 & 75.30 & 55.41 & 44.44 & 17.53 & 26.32 \\
        \midrule
        D & $\checkmark$ & SigLIP & - & Und. Only & 13,000 & 85.03 & 69.10 & 54.23 & - & - \\
        E & $\checkmark$ & - & VAE$^\dagger$+ConvNeXt  & Gen. Only & 37,000 & - & - & - & 16.69 & 26.89 \\
        \midrule
        \textbf{F} & $\checkmark$ & SigLIP & VAE$^\dagger$+ConvNeXt & Unified & 50,000 & 84.73 & 69.20 & 54.83 & 17.61 & 26.40 \\
        \bottomrule
    \end{tabular}
    \label{tab:exp-ablate}
\end{table}

We conduct comprehensive ablation studies to validate the effectiveness of our key design choices. For computational efficiency, all ablation experiments are performed on $256 \times 256$ resolution images\footnote{The understanding encoders in the $256\times256$-based ablation studies is also SigLIP-Large-Patch/16 which is pre-trained on $256\times256$ images.}. All models are trained on our unified pre-training dataset for $50,000$ iterations, except for the understanding-only and generation-only variants, which are trained for proportionally fewer iterations based on their respective data ratios in the pre-training phase. The quantitative results of these ablation studies are presented in Tab.~\ref{tab:exp-ablate}.

\noindent \textbf{Impact of Representation Alignment.}
The comparison between Exp. A and F demonstrates the significant benefits of incorporating representation alignment regularization~\citep{2024REPA} during training. Specifically, models trained with representation alignment show notably lower FID scores on MJHQ dataset and higher CLIP scores, indicating simultaneous improvements in both image quality and semantic alignment. Importantly, our architecture differs from previous studies~\citep{2023DiT, 2024sit} examined in~\citep{2024REPA} due to our incorporation of LLM and an additional skip connection between $g_{enc}$ and $g_{dec}$. The effectiveness of representation alignment in our modified architecture suggests its broad applicability and generalization capability across different network structures.

\begin{figure}[t]
    \centering
    \includegraphics[width=0.97\linewidth]{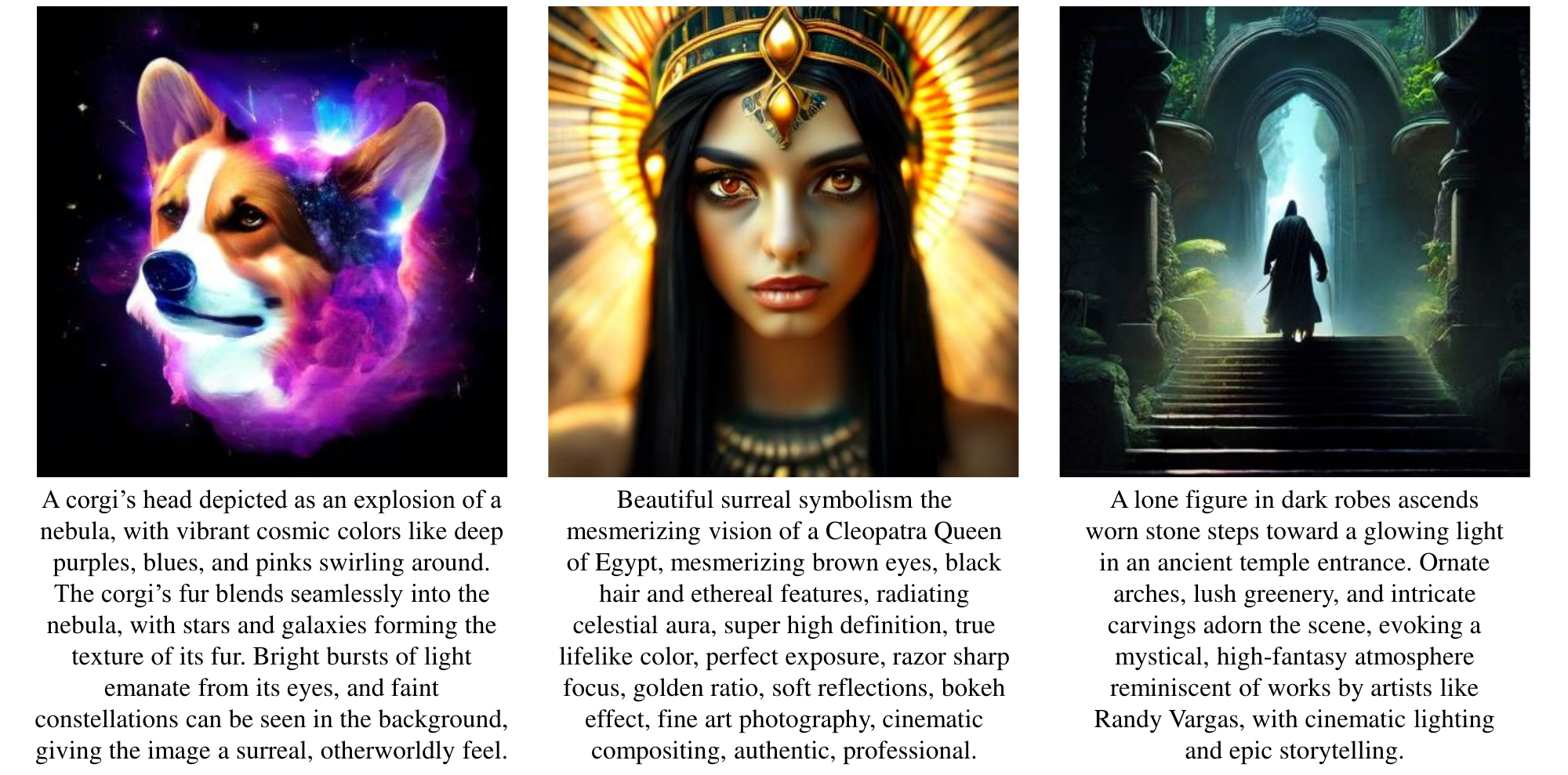}
    \caption{\textbf{Image generation results of \ours.} 
    Our model can generate high-quality images that are semantically consistent with text prompts.
    }
    \label{fig:exp-main_gen}
\end{figure}

\noindent \textbf{Impact of Decoupling Visual Encoders.} e efficacy of using powerful pre-trained visual encoders in multimodal understanding.
The comparison among Exp. B, C, and F demonstrates the advantages of using separate visual encoders for understanding and generation tasks. In Exp. B, following a design similar to Transfusion~\citep{2024Transfusion}, we implement shared ConvNeXt blocks in the SDXL-VAE latent space for both understanding and generation encoders. Exp. C employs separate encoders with identical architectures and initialization parameters, but trained independently. The performance differences between these configurations validate the necessity of decoupled visual encoders in improving our unified model's capabilities. Moreover, the superior results in Exp. C and F highlight the benefits of leveraging pre-trained semantic visual encoders for multimodal understanding tasks.

\noindent \textbf{Fair Comparison with Understanding / Generation-Only Models.}
To establish meaningful benchmarks, we evaluate task-specific models trained under identical conditions - using the same pre-training dataset, infrastructure, and hyperparameters. Exp. D and E represent these specialized models, trained with data volumes matching the unified models in Tab.~\ref{tab:exp-ablate}. The minimal performance gap between Exp. F and these task-specific baselines demonstrates that our unified framework successfully integrates understanding and generation capabilities without significant compromise in either task's performance.

\begin{figure}[t]
    \centering
    \includegraphics[width=\linewidth]{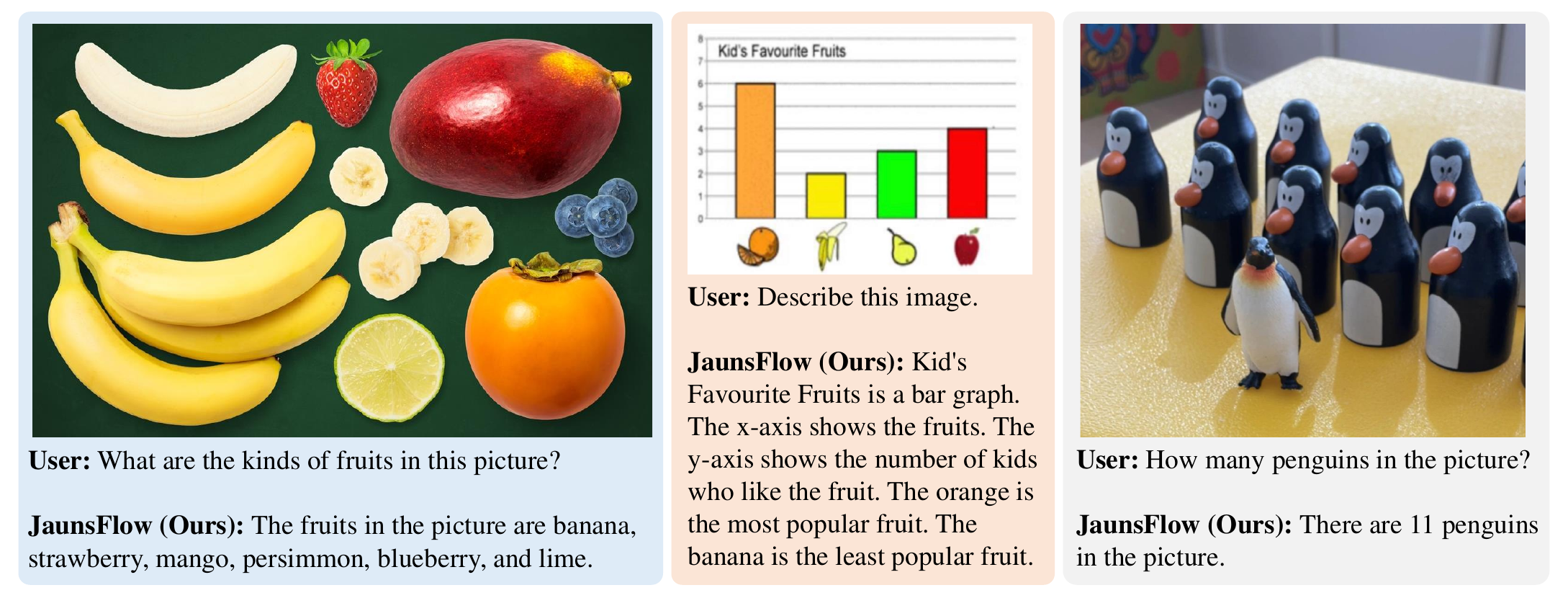}
    \caption{\textbf{Visual Understanding with \ours.} Our model effectively handles various visual understanding tasks, such as question answering, plot interpretation and object counting.}
    \label{fig:exp-main_und}
\end{figure}

\subsection{Qualitative Results}

We present qualitative evaluations of our method for both image generation and understanding tasks. Fig.~\ref{fig:teaser_generation} and Fig.~\ref{fig:exp-main_gen} showcases the image generation capabilities of \ours.
These results demonstrate both the high visual quality of our generated images and our framework's ability to faithfully follow diverse instructions.
For multimodal understanding, Fig.~\ref{fig:exp-main_und} presents example conversations that show our model's understanding capabilities across various scenarios. These interactions demonstrate the model's ability to understand and reason about visual content in natural language dialogues.
Additional qualitative examples showcasing the versatility and effectiveness of \ours~are provided in the appendix.

\section{Conclusion}

We present \ours, a unified framework that successfully harmonizes autoregressive and rectified flow models for multimodal understanding and generation tasks. Our extensive experiments demonstrate that this unification achieves comparable performance to task-specific models. The successful integration of these fundamentally different model architectures not only addresses current challenges in multimodal learning but also opens new possibilities for future research in training unified models.

\clearpage

\bibliographystyle{abbrvnat}
\bibliography{main}

\clearpage

\appendix
\section*{Appendix}

\setcounter{table}{0}
\renewcommand\thetable{\arabic{table}}
\setcounter{figure}{0}
\renewcommand\thefigure{\arabic{figure}}

\section{Performance Analysis of 256 Resolution Model}
We trained our model at two resolutions: $256 \times 256$ and $384 \times 384$. 
The main paper presents results from the $384 \times 384$ model as our primary results.
Here, we provide a comprehensive evaluation of the $256 \times 256$ model's performance.
The visual understanding performances are presented in Tab.~\ref{tab:supp-com}. 
The generation capabilities are evaluated using GenEval~\cite{2024Geneval}, DPG-Benchmark~\cite{2024DPGBench}, and MJHQ FID-30k~\cite{2024PG2.5}, with results shown in Tab.~\ref{tab:supp-geneval} and~\ref{tab:supp-dpg-fid}.
\begin{table}[h]
    \centering
    \setlength{\tabcolsep}{2.2pt}
    \renewcommand{\arraystretch}{1.2}
    \scriptsize
    \caption{Results on visual understanding tasks.}
    \label{tab:supp-com}
    \begin{tabular}{lcccccccc}
        \toprule
        \textbf{Model} & \textbf{LLM Params} & \textbf{POPE$\uparrow$} & \textbf{MME-P$\uparrow$} & \textbf{MMB$_{dev}\uparrow$} & \textbf{SEED$\uparrow$} & \textbf{VQAv2$_{test}$$\uparrow$} & \textbf{GQA$\uparrow$} & \textbf{MM-Vet$\uparrow$} 
        \\
        \midrule
        \textbf{\ours~256} & 1.3B & 85.3 & 1203.0 & 71.9 & 67.6 & 76.3 & 58.4 & 27.4 
        \\
        \textbf{\ours~384} & 1.3B & 88.0 & 1333.1 & 74.9 & 70.5 & 79.8 & 60.3 & 30.9 
        \\
        \bottomrule
    \end{tabular}
\end{table}

\begin{table}[h]
    \centering
    \setlength{\tabcolsep}{4pt}
    \renewcommand{\arraystretch}{1.2}
    \scriptsize
    \caption{Results on GenEval~\cite{2024Geneval}.}
    \label{tab:supp-geneval}
    \begin{tabular}{lcccccccc}
        \toprule
        \textbf{Method} & \textbf{LLM Params} & \textbf{Single Obj.} & \textbf{Two Obj.} & \textbf{Count.} & \textbf{Colors} & \textbf{Pos.} & \textbf{Color Attri.} & \textbf{Overall$\uparrow$} \\
        \midrule
        \textbf{\ours~256} & 1.3B & 0.98 & 0.73 & 0.54 & 0.83 & 0.63 & 0.53 & 0.70 \\
        \textbf{\ours~384} & 1.3B & 0.97 & 0.59 & 0.45 & 0.83 & 0.53 & 0.42 & 0.63 \\
        \bottomrule
    \end{tabular}
\end{table}

\begin{table}[h]
    \centering
    \renewcommand{\arraystretch}{1.2}
    \scriptsize
    \caption{Results on DPG-Bench~\cite{2024DPGBench} and MJHQ FID-30k~\cite{2024PG2.5}.}
    \begin{tabular}{lccccccc}
        \toprule
        \multirow{2}{*}{\textbf{Method}} & \multicolumn{6}{c}{\textbf{DPG-Bench$\uparrow$}} & \multirow{2}{*}{\textbf{MJHQ FID-30k$\downarrow$}} \\
        & \textbf{Global} & \textbf{Entity} & \textbf{Attribute} & \textbf{Relation} & \textbf{Other} & \textbf{Overall} & 
        \\
        \midrule
        \textbf{\ours~256} & 91.20 & 88.83 & 88.00 & 87.60 & 89.53 & 81.23 & 12.70 \\
        \textbf{\ours~384} & 87.03 & 87.31 & 87.39 & 89.79 & 88.10 & 80.09 & 9.51 \\
        \bottomrule
    \end{tabular}
    \label{tab:supp-dpg-fid}
\end{table}

As expected, the $256 \times 256$ model shows slightly lower performance compared to the $384 \times 384$ model on visual understanding metrics due to its reduced resolution. 
Interestingly, however, the $256 \times 256$ model outperforms its higher-resolution counterpart on GenEval and DPG-Bench - benchmarks specifically designed to evaluate instruction following capabilities and semantic accuracy. This superior performance on semantic tasks can be attributed to the model's better control over lower-resolution images, where reduced visual complexity allows for more precise semantic manipulation.

\section{Details of the Datasets}
The datasets used in the pre-training stage for understanding include DetailedCaption~\cite{2023DetailedCaption}, SAM~\cite{2023SAM}, arXivQA\cite{2024ArxivQA}, DenseFusion-1M~\cite{2024DenseFusion}, MMSci\cite{2024MMSci}, PixelProse~\cite{2024PixelProse}, re-captioned LAION-Aesthetics~\cite{2022LA}, re-captioned Open Images V4~\cite{2020OpenImgsv4}, ShareGPT4V~\cite{2023Sharegpt4v}, WikiHow~\cite{2018wikihow} and WIT~\cite{2021wit}. The datasets used in the pre-training stage for generation include re-captioned LAION-Aesthetics~\cite{2022LA}, DALL-E 3 1M~\cite{2024DALLE3-Img}, SAM~\cite{2023SAM}, Open Images V4~\cite{2020OpenImgsv4}, Megalith-10M~\cite{2024Megalith}, YFCC-15M~\cite{2024Yfcc}, PixelProse\cite{2024PixelProse} and JourneyDB~\cite{2024JDB}.

\section{Analysis of CFG Factor and Sampling Steps}
\begin{figure}[t]
\begin{center}
    \subfigure[Results of varying CFG Factors]
    {
        \includegraphics[width=0.48\linewidth]{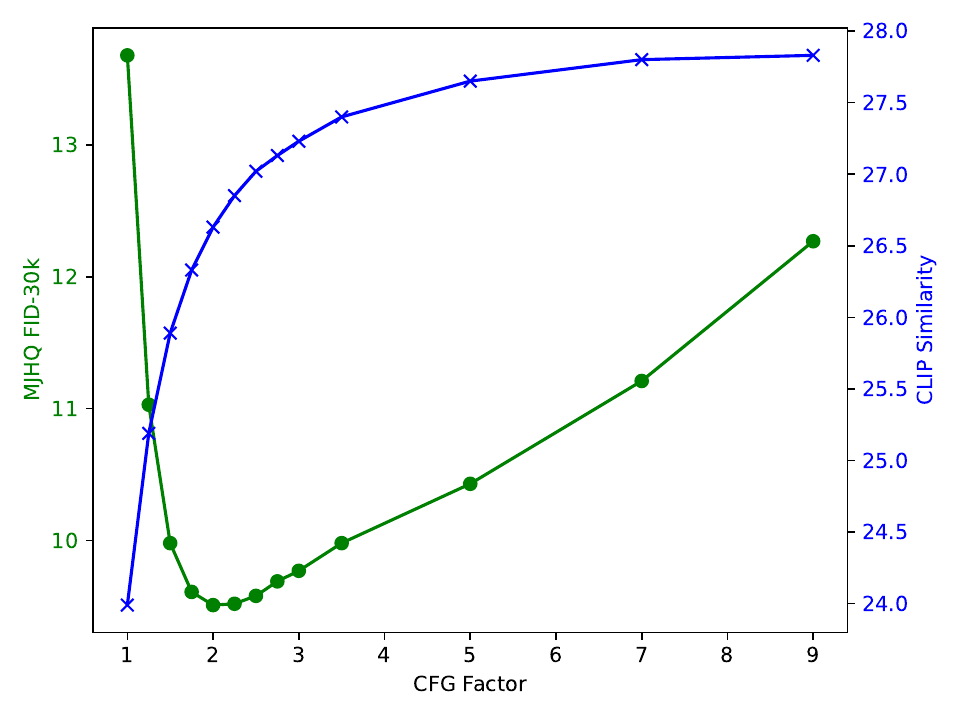}
        \label{fig:supp-sweep-cfg}
    }
    \hfill
    \subfigure[Results of Varying Numbers of Sampling Steps]
    {
        \includegraphics[width=0.48\linewidth]{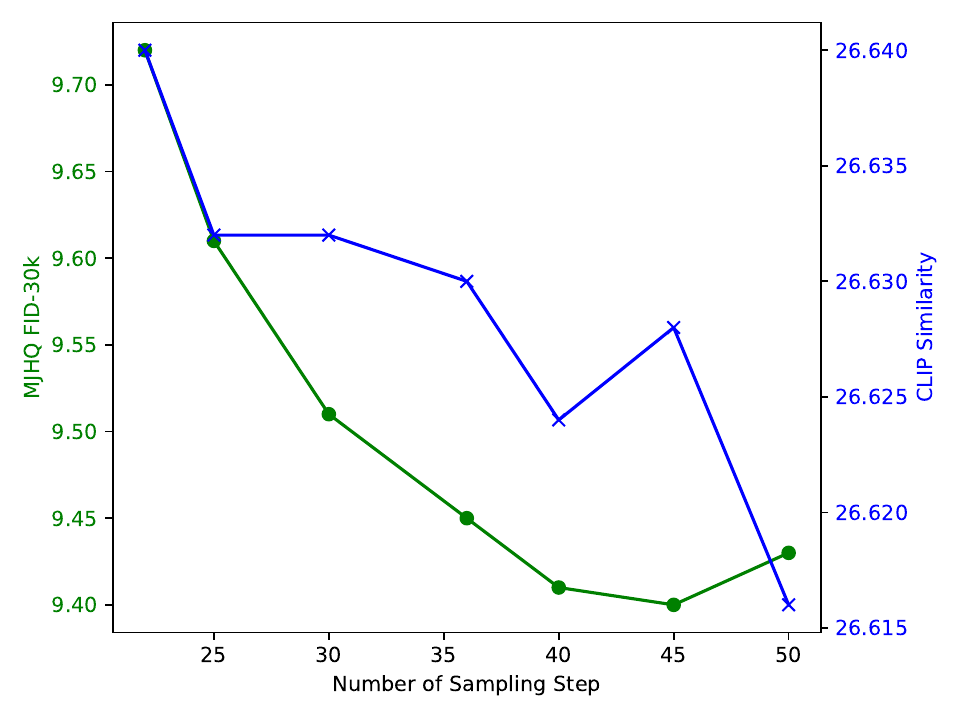}
        \label{fig:supp-sweep-step}
    }
\end{center}
\caption{\small
\textbf{Results of varying CFG factors and numbers of sampling steps}. In Fig.~(a), the number of sampling steps is set to 30. In Fig.~(b), the CFG factor is set to 2.
}
\vspace{-2pt}
\label{fig:supp-sweep}
\end{figure}

\begin{figure}
    \centering
    \includegraphics[width=0.5\linewidth]{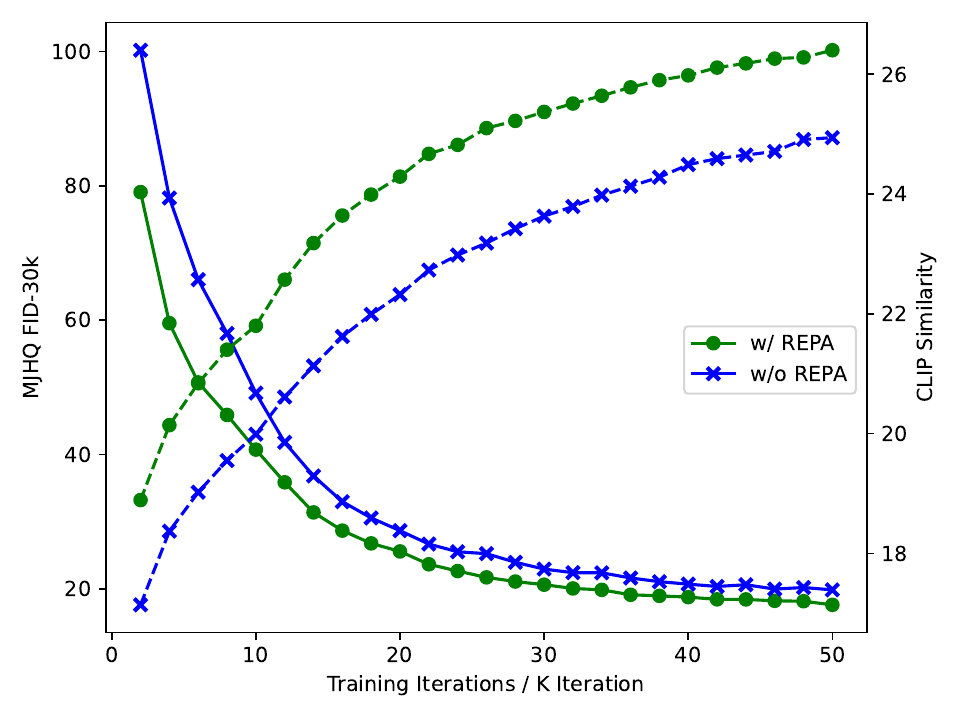}
    \caption{The FID and CLIP similarity during the first 50,000 iterations.}
    \label{fig:supp-repa}
\end{figure}

We investigate the impact of two key generation parameters: the Classifier-Free Guidance (CFG) factor and the number of sampling steps. While our main results use $w=2$ for CFG and 30 sampling steps to calculate FID, here we present a comprehensive analysis of these hyperparameters.
Fig.~\ref{fig:supp-sweep-cfg} shows the effect of varying CFG factors while maintaining 30 sampling steps. The results reveal an optimal CFG value for FID scores, while CLIP \cite{2021CLIP} similarity continues to improve with increasing CFG values, consistent with findings from previous work~\cite{2023SDXL}.
Fig.~\ref{fig:supp-sweep-step} demonstrates the impact of different sampling steps while maintaining a CFG factor of 2. The number of sampling steps shows relatively minor influence on performance. Our choice of 30 steps in the main paper represents a balance between generation quality and computational efficiency.

\section{Details of REPA Ablation}
We provide the FID and CLIP similarity of the first 50,000 training iterations of the pre-train stage in Fig.~\ref{fig:supp-repa} with and without representation alignment regularization. The gap between the two models demonstrates the benefits of using representation alignment regularization.

\section{Additional Qualitative Results}

\begin{figure}[h]
    \centering
    \includegraphics[width=0.79\linewidth]{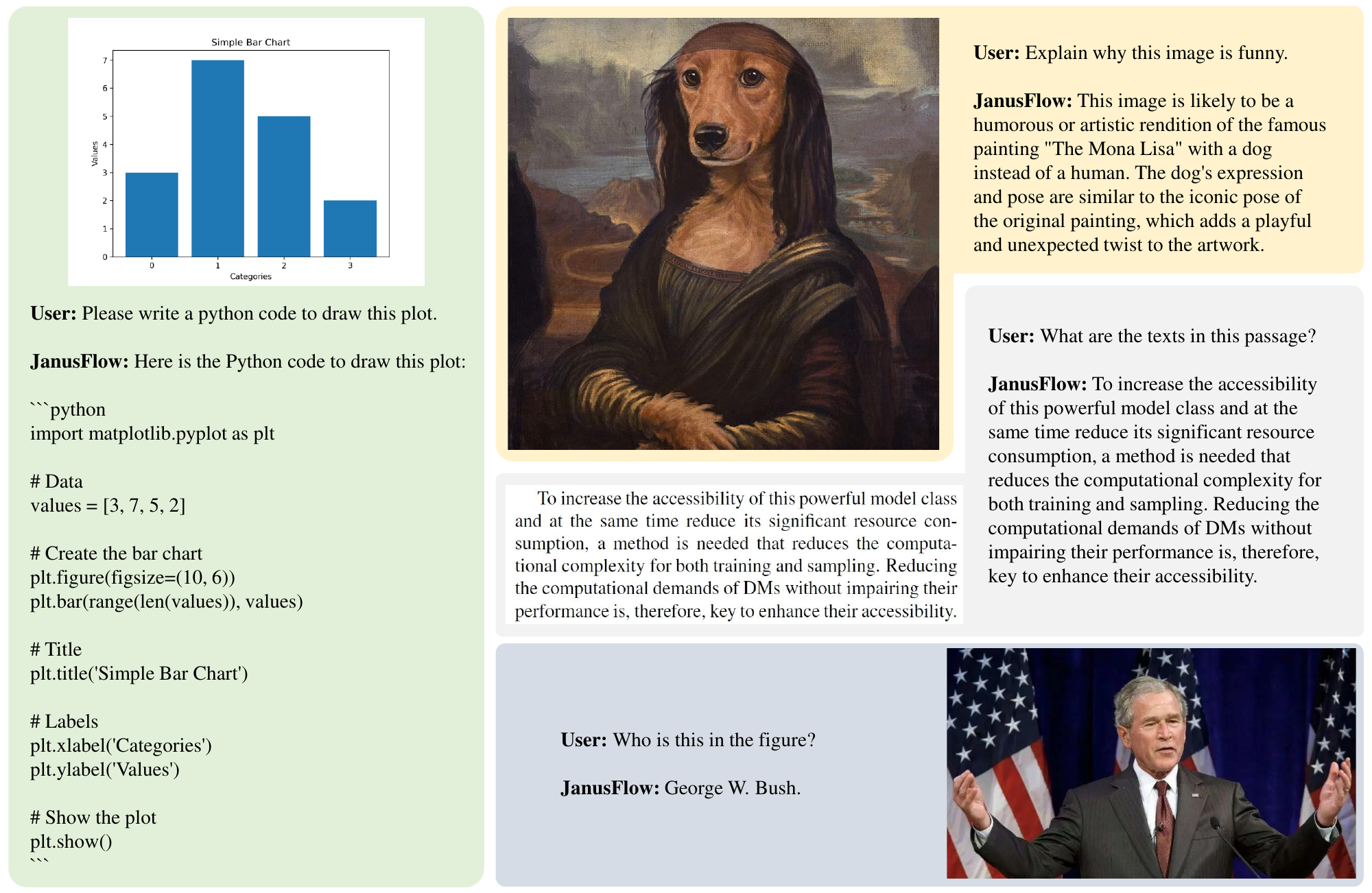}
    \caption{More multimodal understanding cases.}
    \label{fig:supp-und}
\end{figure}

\begin{figure}
    \centering
    \includegraphics[width=\linewidth]{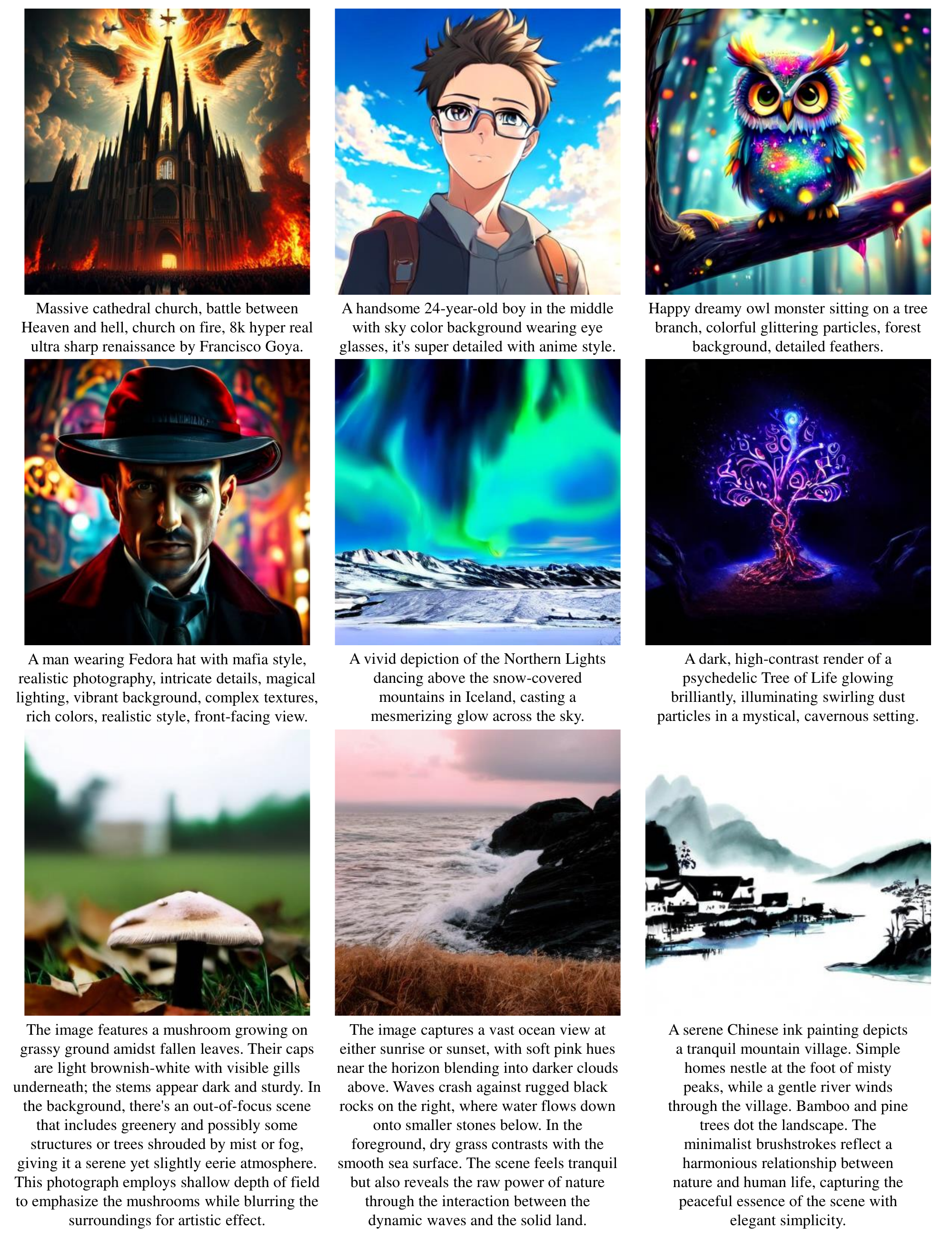}
    \caption{More text-to-image generation results.}
    \label{fig:supp-gen}
\end{figure}

\noindent Additional qualitative examples for both understanding and generation tasks are presented in Fig.~\ref{fig:supp-und} and Fig.~\ref{fig:supp-gen}, respectively. The understanding examples demonstrate \ours's diverse capabilities, including code generation, person identification, character recognition, and visual reasoning. For image generation, our model exhibits strong performance in both visual quality and semantic alignment with input prompts.

\end{CJK*}
\end{document}